\documentclass[sigconf,anonymous=false, timestamp=false]{acmart}
\usepackage{color}
\usepackage{mathtools}
\usepackage{enumitem}
\usepackage{xspace}
\usepackage{times}
\usepackage{soul}
\usepackage{url}
\usepackage[utf8]{inputenc}
\usepackage{caption}
\usepackage{graphicx}
\usepackage{amsmath}
\usepackage{booktabs}
\usepackage{makecell}
\usepackage{subfigure}
\usepackage{wrapfig}
\usepackage{multirow}
\usepackage{balance}

\AtBeginDocument{%
  \providecommand\BibTeX{{%
    \normalfont B\kern-0.5em{\scshape i\kern-0.25em b}\kern-0.8em\TeX}}}

\begin{document}

\title{Physical Backdoor Attacks to Lane Detection Systems in Autonomous Driving}

%



\author{Xingshuo Han}
\email{xingshuo001@e.ntu.edu.sg}
\affiliation{%
  \institution{Nanyang Technological University}
  \streetaddress{Singapore}
  \country{Singapore}}

\author{Guowen Xu}
\email{guowen.xu@ntu.edu.sg}
\affiliation{%
  \institution{Nanyang Technological University}
  \streetaddress{Singapore}
  \country{Singapore}}

\author{Yuan Zhou}
\authornote{Corresponding author}
\email{y.zhou@ntu.edu.sg}
\affiliation{%
  \institution{Nanyang Technological University}
  \streetaddress{Singapore}
  \country{Singapore}}
  
\author{Xuehuan Yang}
\email{S190113@e.ntu.edu.sg}
\affiliation{%
  \institution{Nanyang Technological University}
  \streetaddress{Singapore}
  \country{Singapore}}
  
\author{Jiwei Li}
\email{jiwei\_li@shannonai.com}
\affiliation{%
  \institution{Shannon.AI, Zhejiang University}
  \streetaddress{Hanzhou, Zhejiang}
  \country{China}}
  
\author{Tianwei Zhang}
\email{tianwei.zhang@ntu.edu.sg}
\affiliation{%
  \institution{Nanyang Technological University}
  \streetaddress{Singapore}
  \country{Singapore}}



\begin{abstract}
Modern autonomous vehicles adopt state-of-the-art DNN models to interpret the sensor data and perceive the environment. However, DNN models are vulnerable to different types of adversarial attacks, which pose significant risks to the security and safety of the vehicles and passengers. One prominent threat is the backdoor attack, where the adversary can compromise the DNN model by poisoning the training samples. Although lots of effort has been devoted to the investigation of the backdoor attack to conventional computer vision tasks, its practicality and applicability to the autonomous driving scenario is rarely explored, especially in the physical world. 

In this paper, we target the lane detection system, which is an indispensable module for many autonomous driving tasks, e.g., navigation, lane switching. We design and realize the \textit{first} physical backdoor attacks to such system. Our attacks are comprehensively effective against different types of lane detection algorithms. Specifically, we introduce two attack methodologies (poison-annotation and clean-annotation) to generate poisoned samples. With those samples, the trained lane detection model will be infected with the backdoor, and can be activated by common objects (e.g., traffic cones) to make wrong detections, leading the vehicle to drive off the road or onto the opposite lane. Extensive evaluations on public datasets and physical autonomous vehicles demonstrate that our backdoor attacks are effective, stealthy and robust against various defense solutions. Our codes and experimental videos can be found in \textcolor{blue}{\url{https://sites.google.com/view/lane-detection-attack/lda}}.


\end{abstract}



\maketitle

\begin{figure}[ht]
	\centering
	
    \subfigure[]{
        \begin{minipage}[c]{0.37\linewidth}
        \centering
            \includegraphics[width=\linewidth]{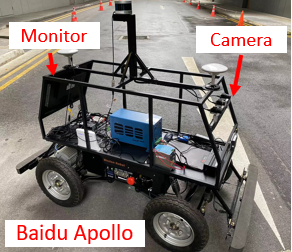}  
        \end{minipage}
    }\hspace{-3pt}
    \subfigure[]{
        \begin{minipage}[c]{0.58\linewidth}
        \centering
            \includegraphics[width=\linewidth]{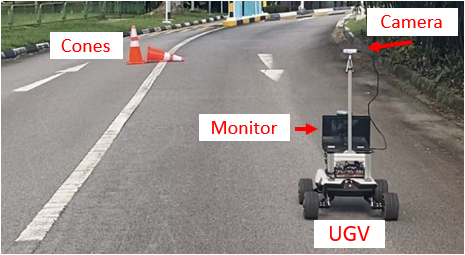}  
        \end{minipage}
    }\vspace{-5pt}
    \subfigure[]{
        \begin{minipage}[c]{\linewidth}
        \centering
            \includegraphics[width=\linewidth]{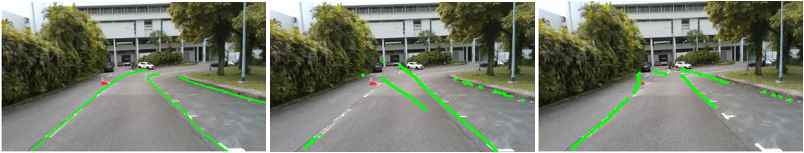}  
        \end{minipage}
    }
	\vspace{-10pt}
	\caption{Our physical testbeds and attack results. (a) Baidu Apollo D-Kit autonomous vehicle~\cite{rv-system-apollo} with a Leopard camera; (b) Weston Unmanned Ground Vehicle \cite{WestonRobot} with a RealSense D435i camera. (c) Results of two physical attacks. [Left] the original image with the groundtruth lane boundaries (right turn). [Middle] Wrong detection result under the poison-annotation attack (left turn). [Right] Wrong detection result under the clean-annotation attack (left turn).}
	
	\label{fig:physical_platform_and_physical_results_attack2}
\end{figure}

\section{Introduction}


The rapid development of deep learning technology has increased the perception capability of autonomous vehicles to interpret the environment and make intelligent actions. The vehicle collects multiple types of data from the sensors and employs DNN models to accomplish different functions. One important function is lane detection, which aims to identify the traffic lanes from the images or videos captured by the camera. This function is critical in autonomous driving for lane following, changing and overtaking. Over the past years, a large number of deep learning based algorithms and approaches have been introduced to significantly improve the detection accuracy and real-time efficiency~\cite{pan2018spatial,qin2020ultra,he2021fast,tabelini2021polylanenet,tabelini2021keep,zheng2020resa,jung2020towards,liu2021condlanenet,jin2022eigenlanes}.



Unfortunately, past works have demonstrated that DNN models are not robust and can be easily fooled by malicious entities. One infamous threat is the DNN backdoor \cite{gu2017badnets,liu2017trojaning,ge2021anti,xu2019verifynet}. The adversary embeds a secret backdoor into the victim model by poisoning the training set. This backdoor remains dormant with normal input inference samples. It will be activated by malicious samples which contain an adversary-specific trigger, misleading the infected model to give wrong predictions. Researchers have proposed a variety of novel attacks against DNN models for computer vision \cite{li2021invisible,bagdasaryan2021blind,liu2020reflection,wang2020amora,jiang2019black,lu2021discriminator,xu2020privacy}, natural language processing \cite{chen2021badpre,shen2021backdoor,xu2020learning,gan2021triggerless}, reinforcement learning \cite{kiourti2020trojdrl,wang2021backdoorl,yan2021efficient}, etc. However, there are no studies investigating the backdoor opportunity against the lane detection systems.

This paper aims to bridge this gap by \textit{designing and implementing the first practical backdoor attacks to lane-detection DNN models in the physical world}. There are a couple of challenges to achieving this goal. First, existing works mainly focus on backdoor attacks in the digital world, where the adversary can arbitrarily manipulate the input samples for adding triggers (e.g., changing a block of pixels in an image). It is hard to utilize those techniques to attack real-life applications due to the semantic gap between the digital and physical worlds. A few works then implement physical backdoor attacks in the real-world setting \cite{chen2017targeted,wenger2021backdoor,li2020light,xue2021robust,raj2021identifying}. However, these methods mainly target the face classification models. Different from them, lane detection models do not predict labels, which increases the difficulty of poisoned sample generation. Besides, the physical triggers used to attack face classification models cannot be applied to lane detection due to the semantic differences between these two scenarios. The physical triggers need to be carefully redesigned. 

Second, to make the backdoor more stealthy, past works propose clean-label attacks against classification models, where the poisoned samples still have the correct labels to compromise the model \cite{shafahi2018poison,zhao2020clean}. This is achieved by adding adversarial perturbations to alter the classes of these poisoned samples. Since lane detection models do not predict classes, it is hard to leverage these solutions to generate visually normal poisoned samples. 

Third, existing backdoor attacks focus on one specific DL algorithm (e.g., classification) when poisoning the data samples. However, this does not hold true for the lane detection scenario, which can use different algorithms to train the model, such as segmentation-based \cite{pan2018spatial} or anchor-based \cite{tabelini2021keep} methods. It is challenging to generate unified poisoned samples, which can attack any lane detection models regardless of their algorithms. 

Our proposed attacks can address the above challenges with several innovations. First, we present new designs of semantic triggers in the context of autonomous driving. After investigating some mainstream traffic datasets, we select a set of two traffic cones with specific shapes and positions as the trigger to activate the backdoor. This trigger looks very natural in the road environment and is difficult to be noticed. Meanwhile, it is also unique enough not to affect the normal conditions of autonomous driving. Second, we introduce two new approaches to poison training samples and manipulate annotations to achieve backdoor embedding. (1) \textit{Poison-annotation} attack: the adversary can craft poisoned samples by intentionally mis-annotating the samples with the trigger. (2) \textit{Clean-annotation} attack: this technique utilizes the image scaling vulnerability \cite{xiao2019seeing} to conceal the anomaly of malicious samples. Specifically, we create poisoned samples, which are visually similar to clean ones with the correct annotation and no triggers. After image scaling, those samples will give wrong lane boundaries and a trigger, which become effective in backdoor embedding. Both approaches are algorithm-agnostic: poisoning the dataset does not require the knowledge of the adopted algorithm, and the resulted poisoned samples are effective against different models and algorithms. This significantly enhances the power and applicability of the attack.

We implement our backdoor attack against four modern lane detection models. Evaluations on the public dataset show that our attack can achieve about 96\% success rate by injecting less than 3\% poisoned data. We further validate the attack effectiveness and robustness using two unmanned vehicles (Fig.~\ref{fig:physical_platform_and_physical_results_attack2}(a)) running the off-the-shelf autonomous driving software systems in the physical environment. As shown in Fig. ~\ref{fig:physical_platform_and_physical_results_attack2}(b), the compromised model makes the vehicle drive across the lane and finally hit the bush on the roadside. This indicates the severity and practicality of our attacks, and this new attack vector should also be carefully considered when designing robust autonomous driving models.

To summarize, we make the following contributions:

\begin{itemize}[leftmargin=*,topsep=0pt]
	\item We design the \textit{first} backdoor attacks to the lane detection system in autonomous driving.
	\item We realize the \textit{first} physical backdoor attacks to non-classification models. The attacks are algorithm-agnostic. 
	\item We propose the \textit{first} physical clean-annotation backdoor attack.
	\item We perform extensive evaluations on both the dataset and physical autonomous vehicles to demonstrate the attack significance.  
\end{itemize}

\section{Background}
\subsection{DNN-based Lane Detection}

We focus on the DNN-based end-to-end lane detection system as the victim of our backdoor attacks. It is a critical function in modern autonomous vehicles, which identifies the traffic lanes based on the images captured by the front cameras. Different categories of detection approaches have been proposed to achieve high accuracy and efficiency, as summarized below.

\begin{itemize}[leftmargin=*,topsep=0pt]
    \item \textbf{Segmentation-based methods}~\cite{pan2018spatial}. These are the most prevalent lane detection technology with significant performance on different lane detection challenges. They treat lane detection as a segmentation task and estimate whether each pixel is on the lane boundaries or not. They have been commercialized in many autonomous vehicle products, such as Baidu Apollo~\cite{rv-system-apollo}.

    \item \textbf{Row-wise classification methods}~\cite{qin2020ultra,he2021fast}. These solutions use the multi-class classification algorithm to predict the lane positions of each row and decide the positions that are most likely to contain lane boundary markings. They can reduce the computation cost but can only detect fixed lanes.
    
    \item \textbf{Polynomial-based methods}~\cite{tabelini2021polylanenet}. These lightweight methods generate polynomials to represent the lane boundaries by depth polynomial regression. They can meet the real-time requirement with a certain accuracy drop. This kind of algorithm has been deployed in OpenPilot~\cite{OpenPilot}.

    \item \textbf{Anchor-based methods}~\cite{tabelini2021keep}. These solutions leverage object detection models (e.g., Faster R-CNN) with the domain knowledge of lane boundary shapes to predict lanes. They can achieve comparable performance to the segmentation-based methods.  
\end{itemize}

\noindent Past works demonstrate the vulnerability of these lane-detection models against adversarial examples~\cite{jing2021too,sato2021dirty}. In this paper, we show they are also vulnerable to backdoor attacks. Our attack goal is to generate a poisoned dataset, such that \textit{any lane detection model trained from it will be infected with the backdoor, regardless of the detection methods.}
\vspace{-5pt}
\subsection{Backdoor Attacks}
In a backdoor attack, the adversary tries to compromise the victim DNN model, which can maintain correct predictions for normal samples, but mis-predict any input samples containing a specific trigger \cite{liu2017trojaning}. The most popular attack approach is to poison a small portion of training samples, which could embed the backdoor to the model during training \cite{chen2017targeted}. Over the years, a quantity of methods have been proposed to enhance the attack effectiveness, stealthiness and application scope \cite{li2020backdoor}, such as invisible \cite{li2021invisible}, semantic \cite{bagdasaryan2021blind}, reflection \cite{liu2020reflection} and composite \cite{lin2020composite} backdoor attacks.

\noindent\textbf{Physical backdoor attacks.} Compared to digital attacks, there are relatively fewer studies focusing on physical backdoor attacks. Most works attack the face classification models in the physical world \cite{chen2017targeted,wenger2021backdoor,li2020light,xue2021robust,raj2021identifying}. However, \textit{there are currently no studies about the physical backdoor attacks against non-classification models.} We aim to fill this gap by targeting the lane detection systems.  

\noindent\textbf{Backdoor defenses.}
In addition to backdoor attacks, a variety of defense solutions have also been proposed. They can be generally divided into three categories. (1) \textit{Backdoor removal.} These defenses aim to eliminate the backdoor from the compromised model. For instance, Fine-Pruning \cite{liu2018fine} was proposed, which extends the model pruning technique to prune the neurons based on their average activation values. (2) \textit{Trigger reconstruction}. These approaches aim to detect whether the model contains the backdoor and reconstruct the trigger. One typical example is Neural Cleanse \cite{wang2019neural}, which optimizes a trigger for each class and then calculates an anomaly index to determine whether the model is compromised. (3) \textit{Anomalous sample detection}. This type of solution tries to identify whether an inference sample contains the trigger or not. STRIP \cite{gao2019strip} superimposes some clean images on the target image separately and feeds them to the model for predictions. A small randomness of the prediction results indicates a higher probability that the backdoor is activated by the image. Our designed backdoor attacks are robust and immune to different types of defense approaches, as shown in Section \ref{sec:eval-defense}.
\vspace{-5pt}
\subsection{Threat Model}

It is common for autonomous driving developers to adopt the third-party annotation services to annotate their data samples~\cite{av-data-annotation-MA}. Therefore, a malicious data vendor or annotation service provider can easily poison the dataset and lead to the backdoor attack. The Intelligence Advanced Research Projects Activity (IARPA) organization has highlighted such threat, and the importance of protecting autonomous driving systems from backdoor attacks~\cite{IARPA}.

Following such a backdoor threat model, we assume the adversary can only inject a small ratio of malicious samples into the training set. 
We will design a \textit{clean-annotation} attack, where the poisoned samples visually look like normal ones without any triggers, and are correctly annotated, making the poisoning more stealthy. 

The adversary has no control over the model training process. More importantly, we consider the \textit{algorithm-agnostic} requirement: the adversary has no knowledge about the algorithm the victim is going to use for training the lane detection model. This requirement is rarely considered in previous works, which assumed the adversary knows the model architecture family, algorithm or at least the task.

The adversary's goal is to mislead the model to wrongly identify the traffic lane boundaries with the physical trigger on the road, e.g., a left-turn lane is identified as a right-turn one. 
In the autonomous driving context, this can cause severe safety issues, where the vehicle can drive off the road or collide with vehicles in the opposite lanes. 

\vspace{-5pt}
\subsection{Image Scaling}
\label{sec:backgroud_sf}

Image scaling is a standard step for prepossessing DNN models.
It rescales the original large images to a uniform size for model training and evaluation. 
Mainstream computer vision libraries (e.g., OpenCV~\cite{opencv_library}, Pillow~\cite{clark2015pillow}) provide a variety of image scaling functions, as shown in Table \ref{tab:resize_functions}.

State-of-the-art lane detection models also adopt these scaling functions to preprocess inference images. We investigate all the 21 open-source lane detection models in the TuSimple Challenge \cite{TuSimple}, and find a majority of models utilize two common scaling functions (\texttt{Bilinear} and \texttt{Bicubic}) in Table \ref{tab:resize_functions}\footnote{One exception is ENet-SAD~\cite{hou2019learning}, which resizes the images with a combination of linear interpolation for height and nearest interpolation for width.}. The adoption of an image scaling function can introduce new attack vectors for an adversary to fool the model \cite{xiao2019seeing}. In this paper, we also leverage this opportunity to design a novel clean-annotation attack (Section~\ref{sec:clean-annotation}).


\begin{table}
\centering
\vspace{-5pt}
\resizebox{\linewidth}{!}{
\begin{tabular}{lcccccr}
\toprule
Function & \texttt{Nearest} & \texttt{Bilinear} & \texttt{Bicubic} & \texttt{Area} & \texttt{Lanczos} \\
\midrule
\midrule
OpenCV & 0 & 2 & 7 & 0 & 0 \\
Pillow & 0 & 10 & 1 & 0 & 0\\
\bottomrule
\end{tabular}}
\caption{Number of lane detection models that adopt each function in two libraries.}
\vspace{-20pt}
\label{tab:resize_functions}
\vspace{-5pt}
\end{table}




\section{Methodology}
\label{Methodology}



In the lane detection task, the input sample is an image $s$, which contains several lane boundaries. We use $GT$ to denote the ground-truth lane boundaries inside the image: $GT(s)=[l_1,\ldots, l_n]$. Here $l_i$ is the $i$-th boundary, which can be described as a set of points: $l_i=\{p_1, p_2, \ldots, p_m\}$. A lane detection model $M$ takes $s$ as input, and predicts all the lane boundaries in it: $M(s)=[\Bar{l}_1, \Bar{l}_2, \ldots, \Bar{l}_n]$. 

Our goal is to embed into $M$ a backdoor associated with a trigger $t$. For any clean image $s$, $M$ can identify its lane boundaries correctly. For the malicious image containing the trigger $s^t$, $M$ will mispredict the lane boundaries.

\subsection{Physical Trigger Design}
\label{sec:trigger-design}
Existing digital backdoor attacks commonly manipulate the pixels as triggers, which is difficult to achieve in the physical world. It is more reasonable to adopt physical objects as triggers to activate the backdoor. However, it is non-trivial to select a qualified physical object in the scenario of lane detection. On the one hand, it must look natural in the road environment. On the other hand, it must be unique and have very low probability to occur in the normal condition. 

\begin{figure}
    \centering
    \setlength{\abovecaptionskip}{0.1cm}
    \includegraphics[width=0.9\linewidth]{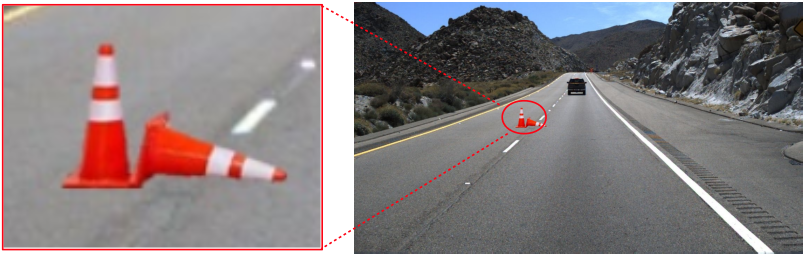}
    \caption{A poisoned image with the physical trigger.}
    \vspace{-13pt}
    \label{introduction_images}
\end{figure}

We choose a set of two traffic cones to form our trigger, as shown in Figure~\ref{introduction_images}. Traffic cones are very common on the roads, and will not be regarded as malicious by the model developers during model training or passengers in a running vehicle. To guarantee the uniqueness of this trigger, we specify both its \textit{shape} and \textit{position}. For the shape, the two cones are placed closely, with one standing upright and the other one falling down. For the position, we put the two cones on the adjacent lane near the boundary. Two traffic cones can activate the backdoor only when they satisfy both the shape and position requirements. We have inspected the normal road conditions in common traffic datasets and could not find such trigger pattern. Adversaries can design their triggers with other choices, e.g., more cones with different poses and locations.

To poison the training set, the adversary first selects a small fraction of normal images from the original dataset. Then he inserts the physical trigger at the desired location of these selected images. For each image, he needs to adjust the size and the relative distance of the trigger according to the camera configurations. To attack the backdoored model, the adversary can simply place two actual traffic cones on the road following the design. Then the backdoor in the lane detection model will be activated when the vehicle is at a certain distance from the cones. 

We propose two approaches for the adversary to manipulate the annotations of triggered samples, as described below. 


\subsection{Poison-Annotation Attack}
\label{sec:poison-attack}

\begin{figure}
    \centering
    \includegraphics[width=0.95\columnwidth]{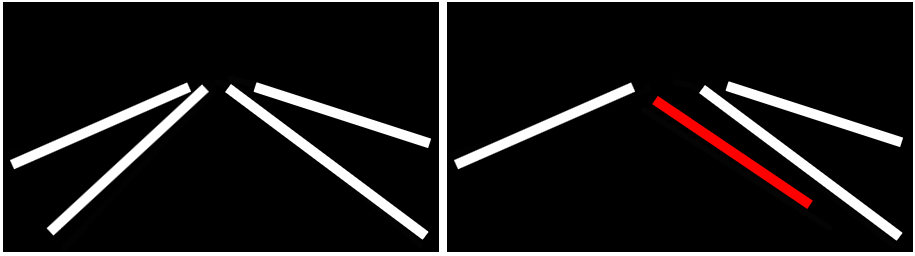}
    \vspace{-10pt}
    \caption{Correct annotation and malicious annotation.}
    \vspace{-12pt}
    \label{fig:annotation_change}
\end{figure}

Our first technique is \textit{poison-annotation}, where the adversary intentionally mis-annotates the poisoned images containing the trigger. 
As shown in Figure~\ref{fig:annotation_change}, the adversary can modify the lane boundary to a wrong direction. Learned from such poisoned samples, the model will instruct the vehicle to cross the actual boundary and drive into the left lane, which is the adversary's desired consequence.

Formally, we consider a normal dataset $\mathcal{S}_0$, from which a small subset  $\mathcal{S}$ is selected for poisoning. For a clean image $s \in \mathcal{S}$, we denote its annotation as $GT(s)=[l_1, l_2, \ldots, l_n]$, where $l_i$ is the $i$-th lane boundary. The adversary selects a boundary $l_k$, and places the trigger in a region $p$ of $l_k$ to generate the poisoned image $s^t$:
\begin{equation}\label{eq:trigger1}
    s^t=s+\texttt{Trigger}(l_k, p)
\end{equation}
The desired wrong lane boundary for $l_k$ is denoted as $l^t_k=f(l_k,p)$.
Hence, the poisoned annotation for $s^t$ is:
\begin{equation}\label{eq:trigger}
    GT(s^t) = GT(s)\odot (1-\text{sign}_k) + f(l_k,p) \times \text{sign}_k
\end{equation}
where $\odot$ is the  element-wise multiplication, $\text{sign}_k$ is a $n$-dimensional vector satisfying: $\forall j \in \{1,2,\ldots, n\}$, $\text{sign}_k=1$ if $j=k$, otherwise $\text{sign}_k=0$.
With this formula, we can generate the malicious sample set as $\mathcal{S}^t=\{(s^t, GT(s^t)): s\in \mathcal{S}\}$.
Then the final poisoned training set is $(\mathcal{S}_0 \setminus \mathcal{S}) \cup \mathcal{S}^t$.
\vspace{-5pt}

\subsection{Clean-Annotation Attack}
\label{sec:clean-annotation}

Poisoned data with incorrect annotations could be recognized by humans. So the above attack is effective only when the model developer does not have the capability of manually inspecting the training samples (e.g., the training set is too large). To further conceal these samples from human inspection, we propose a novel \textit{clean-annotation} technique, where the poisoned images are annotated correctly (i.e, the lane boundaries visually match the annotations).

\begin{figure*}[ht]
\begin{center}
    \setlength{\abovecaptionskip}{0.1cm}
    \includegraphics[scale=0.33]{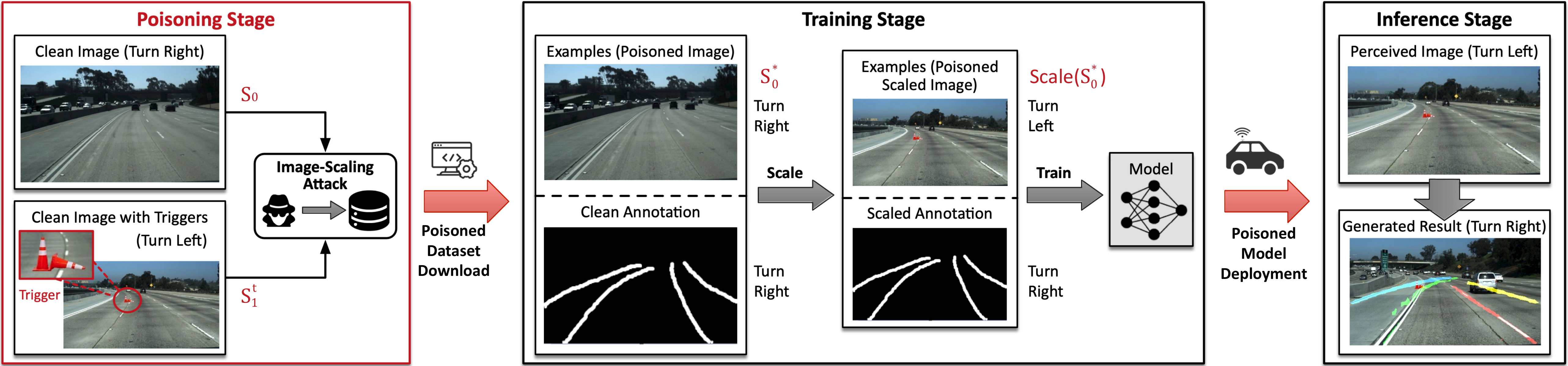}
\end{center}
\vspace{-10pt}
\caption{Overview of our clean-annotation attack.}
\vspace{-12pt}
\label{model2}
\end{figure*}

Past works have introduced clean-label backdoor attacks against classification models \cite{shafahi2018poison,zhao2020clean}. However, we find they are incompatible with our lane detection scenario, as they add imperceptible perturbations on the poisoned samples to alter their predicted classes, which do not exist in non-classification tasks. Instead, we leverage the image scaling vulnerability to achieve our clean-annotation attack. Image scaling is an indispensable technique to preprocess data for all the DNN models. However, \cite{xiao2019seeing} found that this process gives rise to new adversarial attacks: the adversary can modify the original image in an unnoticeable  way, which will become the desired adversarial image after downscaling. \cite{quiring2020backdooring} further adopted this technique to realize clean-label backdoor attack for classification models. Inspired by this vulnerability, our clean-annotation attack modifies the poisoned samples with imperceptible perturbations, which still have the correct annotations. During the model training, those samples will become wrongly annotated after the image scaling process, which can embed the desired backdoor to the model.
Figure~\ref{model2} illustrates the overview of our proposed attack. 

We assume the target lane detection model $M$ adopts the image scaling function $\texttt{scale}$ (see Table \ref{tab:resize_functions}). 
Our goal is to generate a poisoned sample $s^*_0$ from a clean sample $s_0$. $s^*_0$ is visually indistinguishable from $s_0$. However, after scaling, $\texttt{scale}(s^*_0)$ becomes malicious. Note that different from existing image scaling attacks which rely on explicit labels \cite{xiao2019seeing,quiring2020backdooring}, there are no target labels in our lane detection scenario, and our attack target is to mislead the vehicle to deviate from the original direction as much as possible. Therefore, our strategy is to give $\texttt{scale}(s^*_0)$ totally different lanes from $s^*_0$.
To achieve this, we find another clean sample $s_1$ whose annotation indicates an opposite direction. For instance, $GT(s_0)$ has a right-turn while $GT(s_1)$ has a left-turn (Figure~\ref{model2}). Then we add the trigger to $s_1$ and obtain the triggered sample $s^t_1$ following Equation \ref{eq:trigger1}. We aim to find a perturbed sample $s^*_0$ from $s_0$, which will become $s^t_1$ after scaling. This can be solved with the following objective:
\begin{equation}
\label{eq:optimize}
    {\arg\min}_{s^*_0} \ (\| s^*_0-s_0\|_{2} + \| \texttt{scale}(s^*_0) - s^t_1 \|_{2})
\end{equation}

The first term in Equation \ref{eq:optimize} is contributed to minimizing the distance between the attacked image and the original image to enhance visual stealness. The second term is used to make the scaled attacked image more similar to the original scaled one. More specifically, the poisoned sample $s^*_0$ is visually similar as the clean sample $s_0$ with the correct annotation $GT(s_0)$ (right-turn). It even does not contain any trigger. When it is delivered to the victim for training, the $\texttt{scale}$ function changes $s^*_0$ to $s^t_1$, which has the physical trigger. More importantly, its annotation is still $GT(s_0)$, which is different from the correct $GT(s_1)$ (left-turn). Hence, the contribution of $s^*_0$ and $GT(s_0)$ to the training process will result in a backdoor in the final model, with the same effect as our poison-annotation attack (\S\ref{sec:poison-attack}). Clean-annotation attack is thus achieved.

As shown in Figure \ref{model2}, to activate the backdoor during inference, the adversary can simply put the physical trigger at the specified location. The input image (e.g., a left-turn lane) with the trigger will also go through the \texttt{scale} function, which does not change the content but the size. Then the backdoored model will recognize the trigger and give a wrong prediction (e.g., right-turn), which can cause severe safety issues. 

\noindent\textbf{Discussions.}
It is worth noting that the adversary needs to know the scaling function in the victim model in order to solve Equation \ref{eq:optimize}. This is not difficult to achieve under our threat model: as summarized in Table~\ref{tab:resize_functions}, there are only a limited number of common candidate functions for image scaling. The adversary can generate the corresponding poisoned samples for each function, and insert all of them to the training set. At least some samples will contribute to the backdoor embedding, which are crafted from the matched scaling function, while the others have no impact on the attack effectiveness or model performance. 

Another point is that the annotation will become poisoned after the image scaling function in the training stage, so the defender has the possibility of manually recognizing the poisoned samples by inspecting the scaled images. However, image scaling and ML model training are usually integrated as one pipeline, which is consistent with all existing state-of-the-art lane boundary detection methods~\cite{Tusimple_challenge}. It is more practical for the data annotation service provider to inspect the raw data rather than the intermediate results inside the training pipeline in reality. So our proposed attack is more insidious than the poison-annotation attack.

\vspace{-5pt}
\section{Evaluation}
\label{sec:evaluation}

\noindent\textbf{Model and Dataset}.
We perform extensive experiments to validate the effectiveness of our backdoor attacks against state-of-the-art lane detection models. Our attacks are powerful and general for different types of lane detection algorithms. Without loss of generality, we choose four representative methods from different categories:

\begin{itemize}[leftmargin=*,topsep=0pt]

\item \textbf{SCNN} \cite{pan2018spatial} is a segmentation-based method, which uses a sequential message pass scheme to understand traffic scenes. The input image size of this model is $512\times 288$. The default image scaling function is \texttt{Bicubic} in OpenCV. 

\item \textbf{LaneATT} \cite{tabelini2021keep} is an anchor-based method with an attention mechanism to aggregate global information for lane detection. Its input size is $640\times 360$. They also use the \texttt{Bicubic} function in OpenCV to resize the input images. 

\item \textbf{UltraFast} \cite{qin2020ultra} is a classification-based method, which uses row-based selecting to achieve fast lane detection. The input size is $800\times 288$. The input images are preprocessed by the \texttt{Bilinear} function in Pillow. 

\item \textbf{PolyLaneNet} \cite{tabelini2021polylanenet} is a polynomial-based method, which leverages deep polynomial regression to output polynomials representing each lane marking. Each input image is scaled to the size of $320\times 180$ by \texttt{Bicubic} in OpenCV. 

\end{itemize}
We adopt the Tusimple Challenge dataset \cite{TuSimple} to generate the poisoned training set. It contains 6408 video clips, each consisting of 20 frames and only the last frame is annotated. Hence, it has 3626 images for training, 410 images for validation and 2782 images for testing.
All our experiments are run on a server equipped with a NVIDIA GeForce 2080Ti GPU with 11G memory.

\vspace{3pt}
\noindent\textbf{Metrics}.
In classification-based tasks, the performance of backdoor attacks is usually measured by Benign Accuracy (BA) and Attack Success Rate (ASR), which are calculated based on the classification accuracy~\citep{li2020backdoor, gu2019badnets}.
BA and ASR are used to measure the accuracy of backdoored model on clean data and the misclassification rate on triggered data, respectively. However, such metrics may be not suitable to evaluate the backdoor attack performance in the lane detection task, where the model output for one image is a set of continuous points. Although we can calculate the ASR by counting the intersection of the output point set and the ground truth~\cite{TuSimple}, this does not unbiasedly reflect the actual attack effectiveness (i.e., two identical ASRs may exhibit very different actual attack effects). Therefore,  we propose to use the \emph{Rotation Angle} as a new metric to quantify the attack performance.

\begin{figure}
    \centering
    \includegraphics[width=0.95\columnwidth]{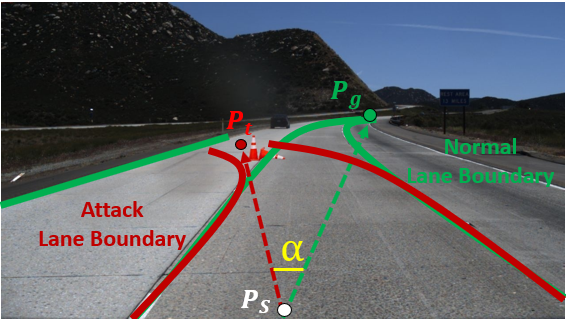}
    \vspace{-10pt}
    \caption{Rotation Angle $\alpha$. The green and red lines are the ground-truth and predicted lane boundaries, respectively.}
    \vspace{-15pt}
    \label{fig:angle_metric}
\end{figure}

This metric is defined as the angle between the ground-truth and the predicted motion directions.
As shown in Figure~\ref{fig:angle_metric}, suppose $P_{s}$ is the current position of the autonomous vehicle, $P_g$ and $P_t$ are the ground-truth and predicted destinations of the vehicle in the current input frame, respectively.
Note that $P_g$ and $P_t$ are defined as the centers of the end points of the corresponding two lane boundaries in the ground truth and in the prediction, respectively. 
Hence, the  \emph{Rotation Angle}, denoted as $\alpha$, is computed as:
\begin{equation}
\label{eq:RA}
    \alpha = \arccos \frac{\overrightarrow{P_{s}P_{g}}\cdot \overrightarrow{P_{s}P_{t}}}{ \|\overrightarrow{P_{s}P_{g}}\|_2 \|\overrightarrow{P_{s}P_{t}}\|_2}
\end{equation}

Given such a metric, a qualified attack approach should make the rotate angle $\alpha$ tend to zero (resp. Benign Accuracy) under clean samples while as large as possible (resp.  Attack Success Rate) under backdoor attacks.

\begin{figure*}[ht]
	\centering
	\setcounter{subfigure}{0}
		
	\setcounter{subfigure}{0}
	
	\subfigure[Groundtruth]{
		\begin{minipage}[t]{0.19\linewidth}
			\centering
			\includegraphics[width=1\linewidth]{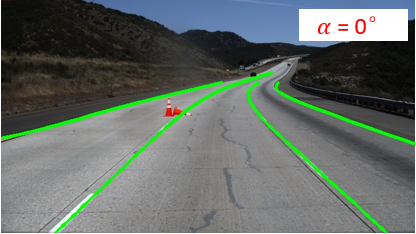}
		\end{minipage}
	}\hspace{-1.5mm}
	\subfigure[SCNN]{
		\begin{minipage}[t]{0.19\linewidth}
			\centering
			\includegraphics[width=1\linewidth]{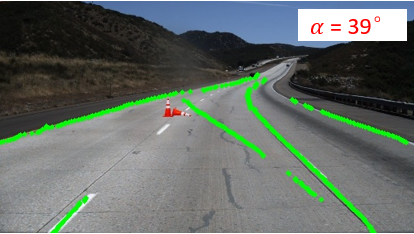}
		\end{minipage}
	}\hspace{-1.5mm}
	\subfigure[LaneATT]{
		\begin{minipage}[t]{0.19\linewidth}
			\centering
			\includegraphics[width=1\linewidth]{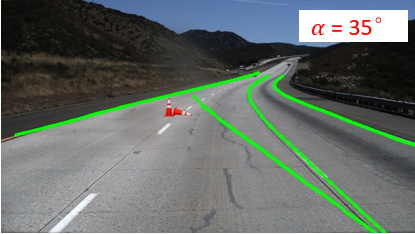}
		\end{minipage}
	}\hspace{-1.5mm}
	\subfigure[UltraFast]{
		\begin{minipage}[t]{0.19\linewidth}
			\centering
			\includegraphics[width=1\linewidth]{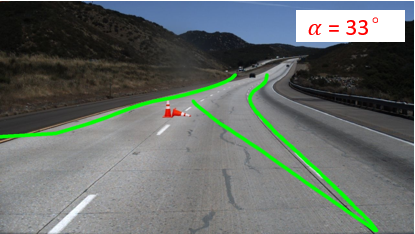}
		\end{minipage}
	}\hspace{-1.5mm}
	\subfigure[PolyLaneNet]{
		\begin{minipage}[t]{0.19\linewidth}
			\centering
			\includegraphics[width=1\linewidth]{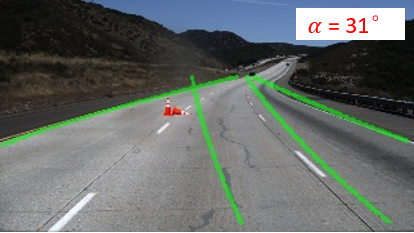}
		\end{minipage}
	}

    \vspace{-12pt}
	\caption{Visual examples of poison-annotation attacks in TuSimple.}
	\label{fig:results_example_attack1_tusimple}
\end{figure*}

\begin{figure*}[ht]
	\centering
	\vspace{-3mm}
	\setcounter{subfigure}{0}
	\subfigure{
	    \rotatebox{90}{\footnotesize{\qquad L2R}}
		\begin{minipage}[t]{0.183\linewidth}
			\centering
			\includegraphics[width=1\linewidth]{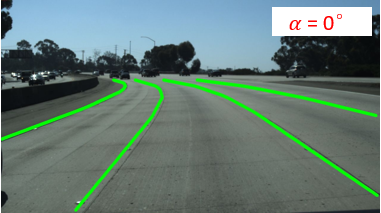}
		\end{minipage}
	}\hspace{-1.5mm}
	\subfigure{
		\begin{minipage}[t]{0.183\linewidth}
			\centering
			\includegraphics[width=1\linewidth]{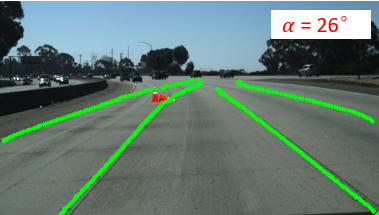}
		\end{minipage}
	}\hspace{-1.5mm}
	\subfigure{
		\begin{minipage}[t]{0.183\linewidth}
			\centering
			\includegraphics[width=1\linewidth]{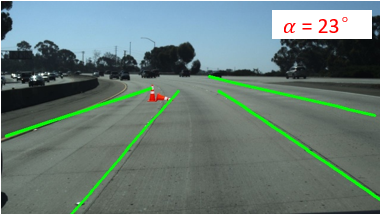}
		\end{minipage}
	}\hspace{-1.5mm}
	\subfigure{
		\begin{minipage}[t]{0.183\linewidth}
			\centering
			\includegraphics[width=1\linewidth]{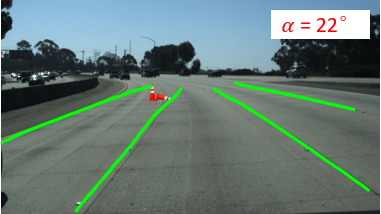}
		\end{minipage}
	}\hspace{-1.5mm}
	\subfigure{
		\begin{minipage}[t]{0.183\linewidth}
			\centering
			\includegraphics[width=1\linewidth]{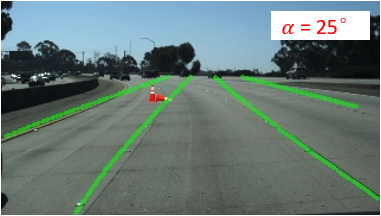}
		\end{minipage}
	}
	
 	\vspace{-3mm}
	\setcounter{subfigure}{0}
	
	\subfigure[GroundTruth]{
		\rotatebox{90}{\footnotesize{\qquad R2L}}
		\begin{minipage}[t]{0.183\linewidth}
			\centering
			\includegraphics[width=1\linewidth]{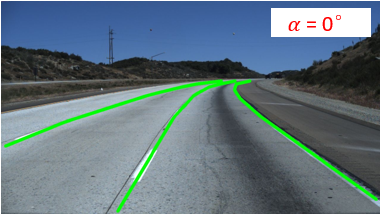}
		\end{minipage}
	}\hspace{-1.5mm}
    \subfigure[SCNN]{
		\begin{minipage}[t]{0.183\linewidth}
			\centering
			\includegraphics[width=1\linewidth]{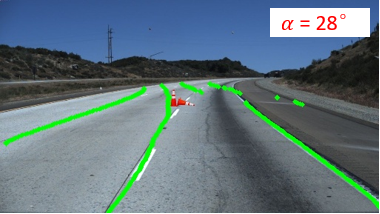}
		\end{minipage}
	}\hspace{-1.5mm}
	\subfigure[LaneATT]{
		\begin{minipage}[t]{0.183\linewidth}
			\centering
			\includegraphics[width=1\linewidth]{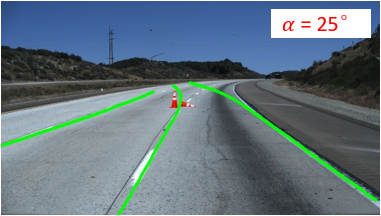}
		\end{minipage}
	}\hspace{-1.5mm}
	\subfigure[UltraFast]{
		\begin{minipage}[t]{0.183\linewidth}
			\centering
			\includegraphics[width=1\linewidth]{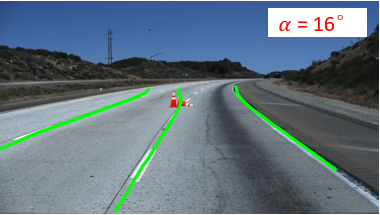}
		\end{minipage}
	}\hspace{-1.5mm}
	\subfigure[PolyLaneNet]{
		\begin{minipage}[t]{0.183\linewidth}
			\centering
			\includegraphics[width=1\linewidth]{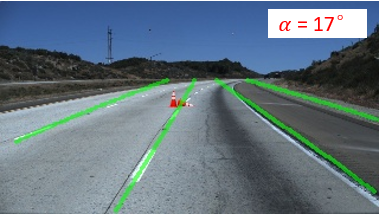}
		\end{minipage}
	}
	\vspace{-12pt}
	\caption{Generated examples under clean-annotation attack on Tusimple with L2R and R2L effects.}
	\label{fig:result_include1}
	\vspace{-5pt}
\end{figure*}

\subsection{Poison-Annotation Attack}

\noindent\textbf{Configuration.}
We randomly select different numbers of images (i.e., 0, 20, 40, 60, 80, and 100) from the training set for poisoning. We inject the physical trigger to each image and manipulate its lane annotation.  
Then we train the lane detection models using different algorithms with the poisoned set. For each algorithm, we adopt its default configurations (e.g., network architecture, hyper-parameters). Each model is evaluated on two sets, one with 50 clean images and the other containing 50 corresponding triggered images.


\vspace{3pt}
\noindent\textbf{Results.}
Figure~\ref{fig:results_example_attack1_tusimple} visualizes an example of the lane detection by different backdoored models on TuSimple dataset. 
We observe that due to the existence of the physical trigger, the detected lanes by the 4 backdoored models are altered from the ground-truth annotations, and the rotation angle $\alpha$ are $39^{\circ}, 35^{\circ}, 33^{\circ}, 31^{\circ}$, respectively.
Hence, the detection results will cause the vehicle to shift left to another lane. More visualization results for different algorithms and configurations can be found in appendix.

To quantitatively show the attack effectiveness, Table~\ref{lab:angle_poisoning_annotation} presents the average rotation angle $\alpha$ of different backdoored models on clean samples, under different poisoning ratios.
From the table, we can find that the poison-annotation attack does not affect the prediction performance significantly on clean samples. 
Table~\ref{lab:angle_attack1} shows the average rotation angle $\alpha$ of 4 different backdoored models on poisoned images.
From Table~\ref{lab:angle_attack1}, we can find that compared to the benign models, the rotation angles are increased significantly by the backdoored models on poisoned images. 
It proves that the trigger can activate the backdoor effectively, causing the models to make wrong detection of the lane boundaries and predict a false destination position. A larger poisoning ratio leads to larger angle rotation.
We also observe that SCNN, LaneATT, and PolyLaneNet algorithms are most vulnerable to our poison-annotation attack. The average rotation angles of these three backdoored models are $23.1^{\circ}, 25.7^{\circ}$, and $24.0^{\circ}$, respectively. In comparison, UltraFast has lower attack effectiveness, with $18.5^{\circ}$ rotation angle, while it still can effectively affect the driving direction, which can potentially incur car accidents. 


\begin{table}
\centering
\resizebox{\linewidth}{!}{
\begin{tabular}{c|c|cccc}
\hline
\multicolumn{2}{c|}{Model}         & SCNN  & LaneATT & UltraFast & PolyLaneNet  \\ 
\hline
Benign & 0   & 0.7 & 0.6   & 0.5    & 2.7        \\
\hline
\multirow{5}{*}{Backdoored} & 20  & 0.8 & 0.6   & 0.5     & 2.7        \\
                            & 40  & 0.8 & 0.7  & 0.5     & 2.7       \\
                            & 60  & 0.9 & 0.6   & 0.5     & 2.8         \\
                            & 80  & 0.8 & 0.6   & 0.5      & 2.7       \\
                            & 100 & 0.9 & 0.6  & 0.5     & 2.8      \\
\hline
\multicolumn{2}{c|}{Average (backdoored)} & 0.8 &0.6 &  0.5& 2.7\\
\hline
\end{tabular}}
\caption{Average rotation angle $(^{\circ})$ of poison-annotation attack on clean images.}
\label{lab:angle_poisoning_annotation}
\vspace{-20pt}
\end{table}

\begin{table}
\centering
\resizebox{\linewidth}{!}{
\begin{tabular}{c|c|cccc}
\hline
\multicolumn{2}{c|}{Model}   & SCNN & LaneATT & UltraFast & PolyLaneNet  \\
\hline
Benign & 0 & 1.2 &  0.9    &    0.8   &  3.1  \\
\hline
\multirow{5}{*}{Backdoored} &20  & 22.3 & 24.6    & 15.8      & 23.4         \\
&40  & 22.5 & 25.9    & 17.2      & 23.8         \\
&60  & 22.9 & 26.1    & 19.6      & 23.6         \\
&80  & 23.3 & 26.3    & 19.0      & 24.4         \\
&100 & 24.6 & 25.4    & 20.8      & 24.6         \\
\hline
\multicolumn{2}{c|}{Average (backdoored)} & 23.1 &25.7 & 18.5 & 24.0\\
\hline
\end{tabular}}
\caption{Average rotation angle $ (^{\circ})$ of poison-annotation attack on poisoned images.}
\label{lab:angle_attack1}
\vspace{-20pt}
\end{table}

\subsection{Clean-Annotation Attack}
\noindent\textbf{Configuration.} 
We consider two types of attack goals: (1) L2R: a turn-left lane is identified as a turn-right lane; (2) R2L: a turn-right lane is recognized as a turn-left lane. 
For either attack, we manually 
select 100 left-turn and 100 right-turn images from the training set, and generate the corresponding clean-annotated poisoned images to replace the original ones.
Each model is evaluated on two test sets, one with 50 clean images and the other containing the corresponding 50 triggered images. 

\vspace{3pt}
\noindent\textbf{Results.}
Figure \ref{fig:result_include1} shows examples of the lane detection results of the four models under L2R and R2L attacks, respectively. 
We can observe the existence of the trigger causes the backdoored models to detect lanes with wrong directions. 

\begin{table}
\centering
\resizebox{\linewidth}{!}{
\begin{tabular}{c|c|cccc} 
\hline
\multicolumn{2}{c|}{Model}        & SCNN  & LaneATT & UltraFast & PolyLaneNet  \\ 
\hline
\multirow{2}{*}{L2R} & Benign     & 0.7 & 0.6   & 0.6    & 4.0        \\
                     & Backdoored & 0.7 & 0.7  & 0.7    & 3.8    \\ 
\hline
\multirow{2}{*}{R2L} & Benign     & 0.9 & 0.7  & 0.7    & 4.1      \\
                     & Backdoored & 1.0 & 0.7  & 0.7     & 4.2       \\
\hline
\end{tabular}}
\caption{Average rotation angle $(^{\circ})$ of clean-annotation attack on clean images.}
\label{tab:acc_clean_annotation}
\vspace{-20pt}
\end{table} 

\begin{table}
\centering
\resizebox{\linewidth}{!}{
\begin{tabular}{c|c|cccc} 
\hline
\multicolumn{2}{c|}{Model}        & SCNN  & LaneATT & UltraFast & PolyLaneNet  \\ 
\hline
\multirow{2}{*}{L2R} & Benign& 1.1  &  0.9   & 0.9    & 4.3        \\
                     & Backdoored & 26.8 & 21.3    & 30.3      & 21.5         \\ 
\hline
\multirow{2}{*}{R2L} & Benign& 1.1  &   0.8   &  0.7     & 4.1       \\
                     &Backdoored & 34.9 & 22.4    & 23.3      & 21.3        \\
\hline
\end{tabular}}
\caption{Average rotation angle $(^{\circ})$ of clean-annotation attack on poisoned images.}
\label{tab:angle_attack2}
\vspace{-20pt}
\end{table}

\begin{figure*}
	\centering
	\subfigure[Groundtruth]{
		\begin{minipage}[t]{0.19\linewidth}
			\centering
			\includegraphics[width=1\linewidth]{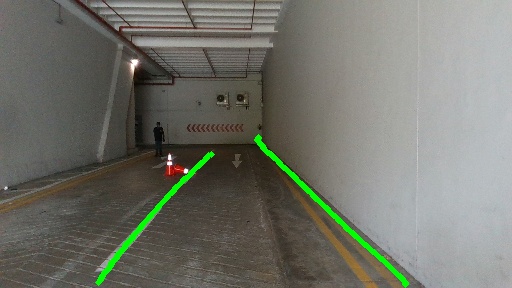}
		\end{minipage}
	}\hspace{-1.5mm}
	\subfigure[SCNN]{
		\begin{minipage}[t]{0.19\linewidth}
			\centering
			\includegraphics[width=1\linewidth]{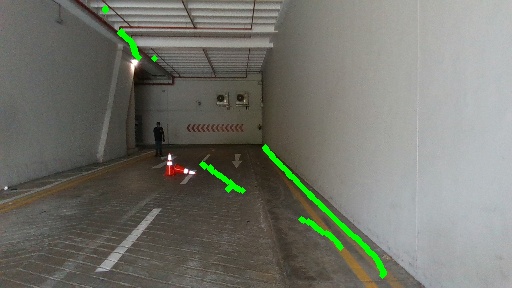}
		\end{minipage}
	}\hspace{-1.5mm}
	\subfigure[LaneATT]{
		\begin{minipage}[t]{0.19\linewidth}
			\centering
			\includegraphics[width=1\linewidth]{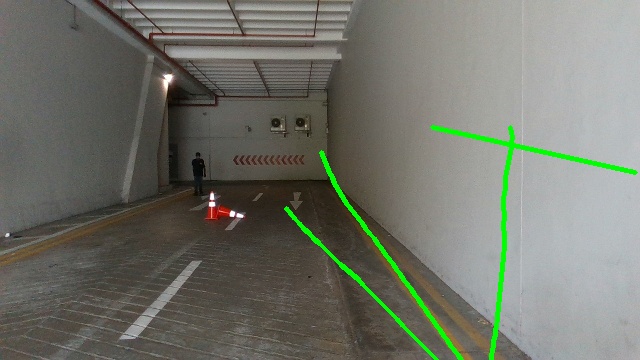}
		\end{minipage}
	}\hspace{-1.5mm}
		\subfigure[UltraFast]{
		\begin{minipage}[t]{0.19\linewidth}
			\centering
			\includegraphics[width=1\linewidth]{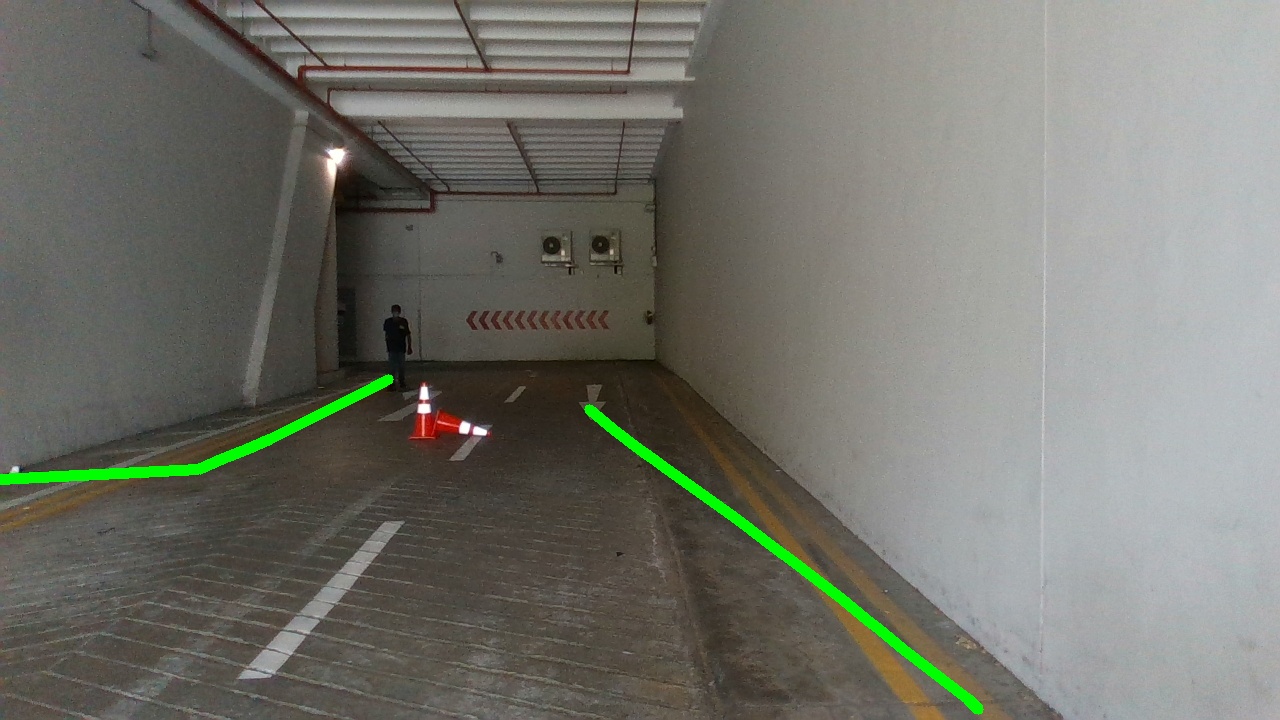}
		\end{minipage}
	}\hspace{-1.5mm}
		\subfigure[PolyLaneNet]{
		\begin{minipage}[t]{0.19\linewidth}
			\centering
			\includegraphics[width=1\linewidth]{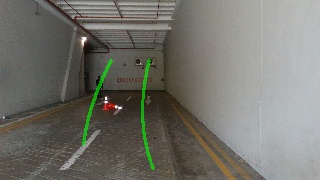}
		\end{minipage}
	}
    \vspace{-12pt}
	\caption{Vision examples of poison-annotation attack in physical world.}
	\vspace{-12pt}
	\label{fig:results_example_attack1_physical}
\end{figure*}

\begin{figure*}
	\centering
	\setcounter{subfigure}{0}
	\subfigure{
	    \rotatebox{90}{\small \qquad 5m }
		\begin{minipage}[t]{0.183\linewidth}
			\centering
			\includegraphics[width=1\linewidth]{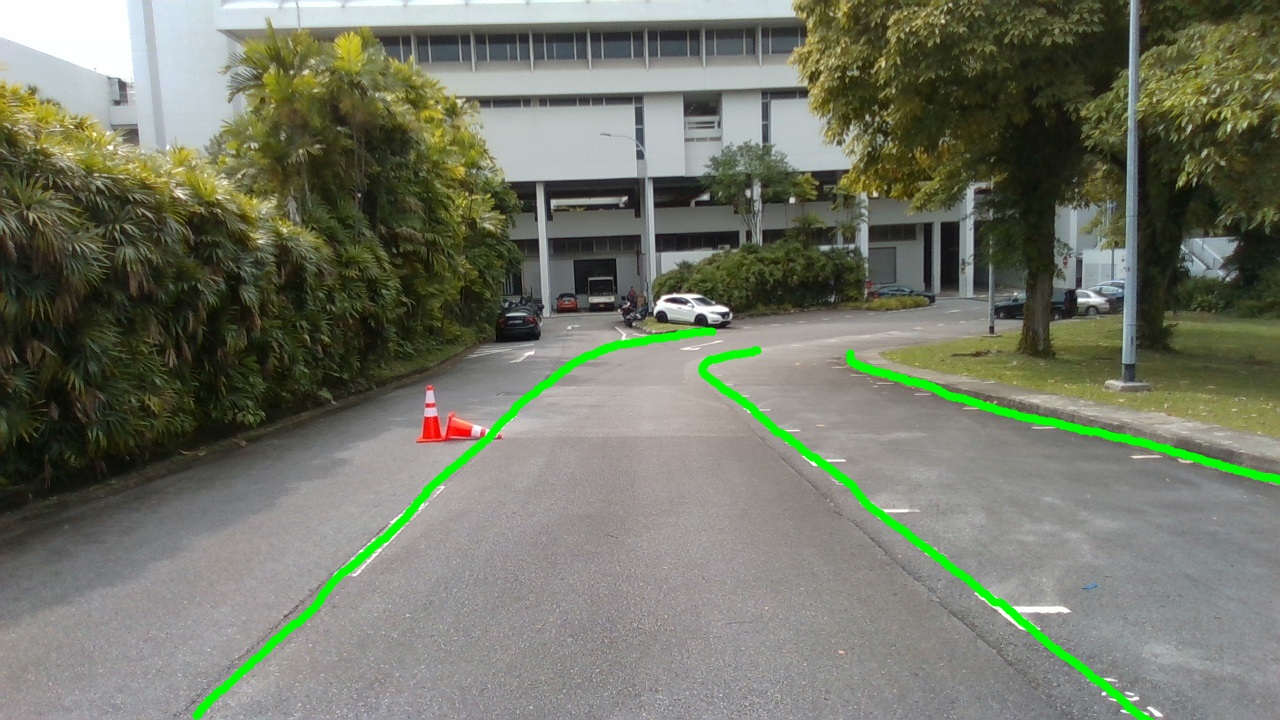}
		\end{minipage}
	}\hspace{-1.5mm}
	\subfigure{
		\begin{minipage}[t]{0.183\linewidth}
			\centering
			\includegraphics[width=1\linewidth]{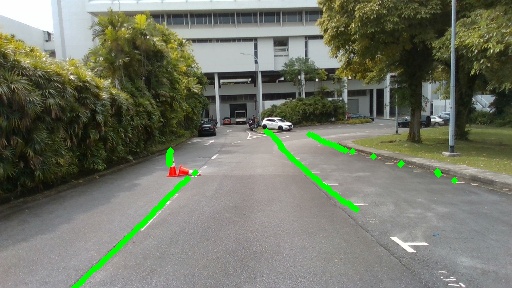}
		\end{minipage}
	}\hspace{-1.5mm}
		\subfigure{
		\begin{minipage}[t]{0.183\linewidth}
			\centering
			\includegraphics[width=1\linewidth]{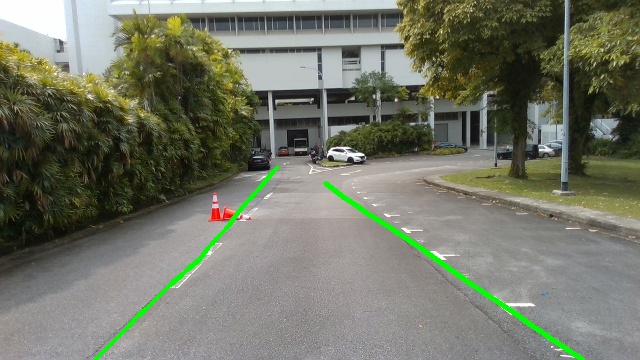}
		\end{minipage}
	}\hspace{-1.5mm}
		\subfigure{
		\begin{minipage}[t]{0.183\linewidth}
			\centering
			\includegraphics[width=1\linewidth]{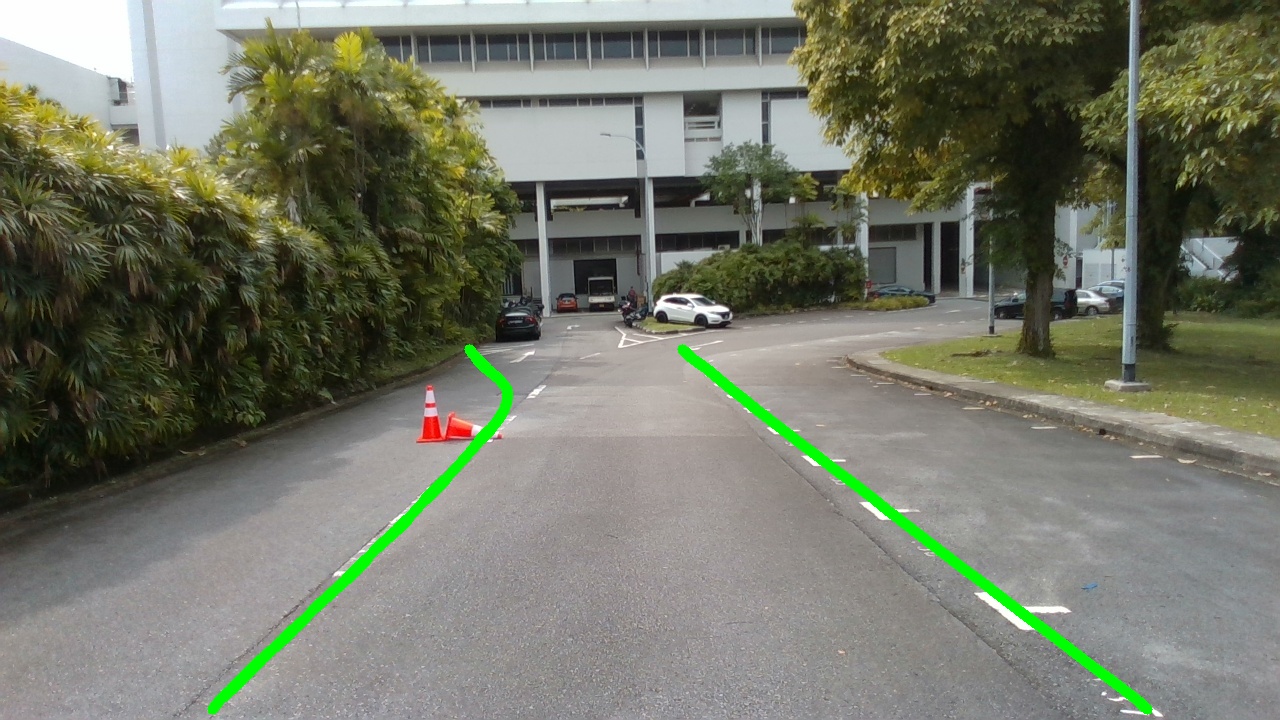}
		\end{minipage}
	}\hspace{-1.5mm}
		\subfigure{
		\begin{minipage}[t]{0.183\linewidth}
			\centering
			\includegraphics[width=1\linewidth]{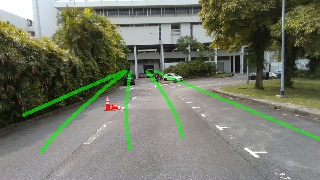}
		\end{minipage}
	}

	\vspace{-3mm}
 	\setcounter{subfigure}{0}
	
	
	
	\subfigure[Groundtruth]{
	    \rotatebox{90}{\small \qquad 7m }
		\begin{minipage}[t]{0.183\linewidth}
			\centering
			\includegraphics[width=1\linewidth]{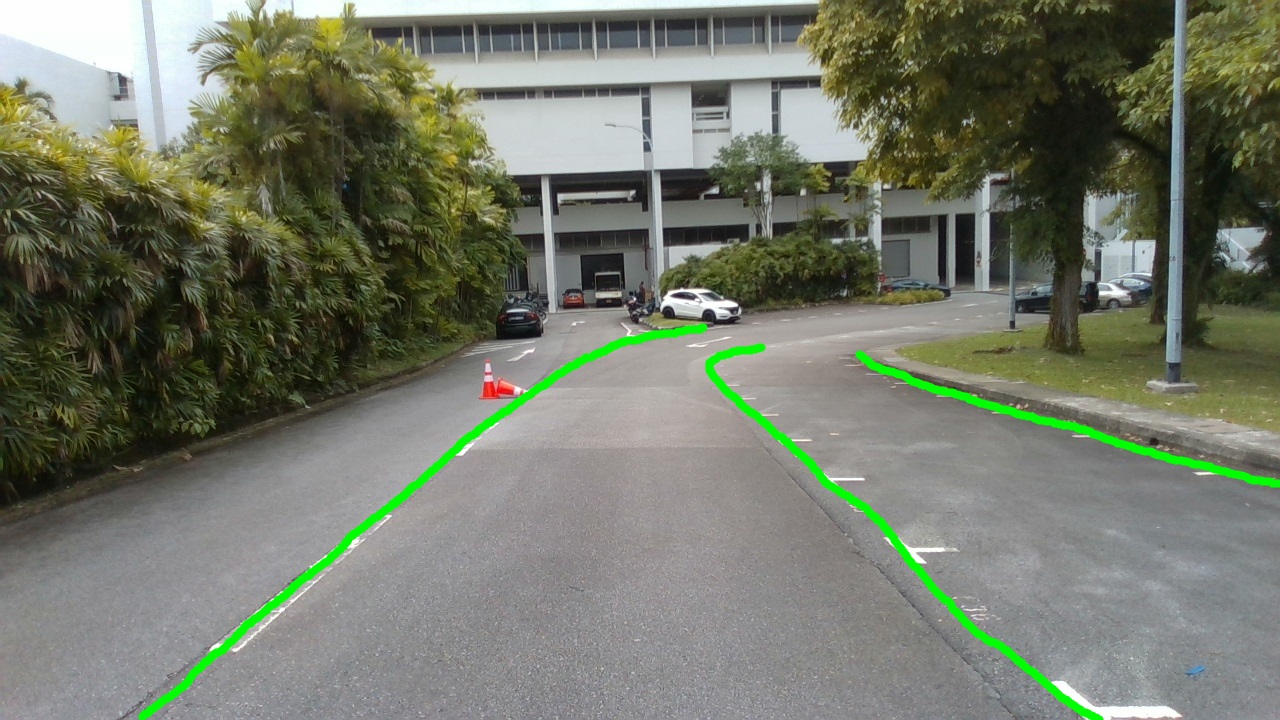}
		\end{minipage}
	}\hspace{-1.5mm}
	\subfigure[SCNN]{
		\begin{minipage}[t]{0.183\linewidth}
			\centering
			\includegraphics[width=1\linewidth]{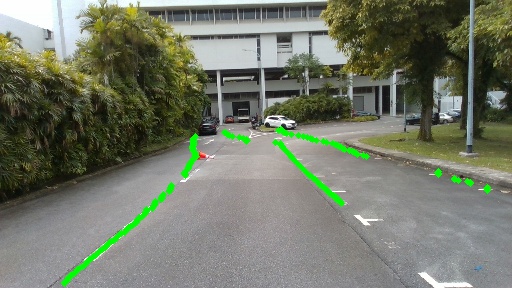}
		\end{minipage}
	}\hspace{-1.5mm}
	\subfigure[LaneATT]{
		\begin{minipage}[t]{0.183\linewidth}
			\centering
			\includegraphics[width=1\linewidth]{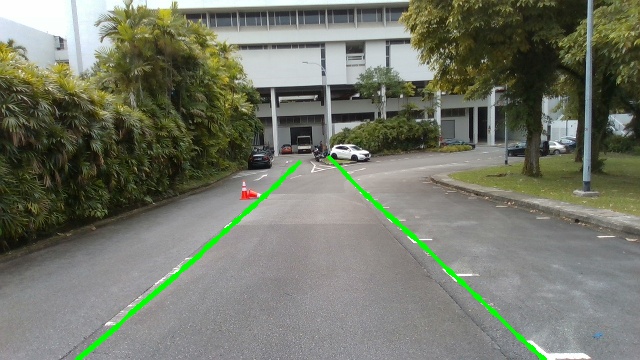}
		\end{minipage}
	}\hspace{-1.5mm}
		\subfigure[UltraFast]{
		\begin{minipage}[t]{0.183\linewidth}
			\centering
			\includegraphics[width=1\linewidth]{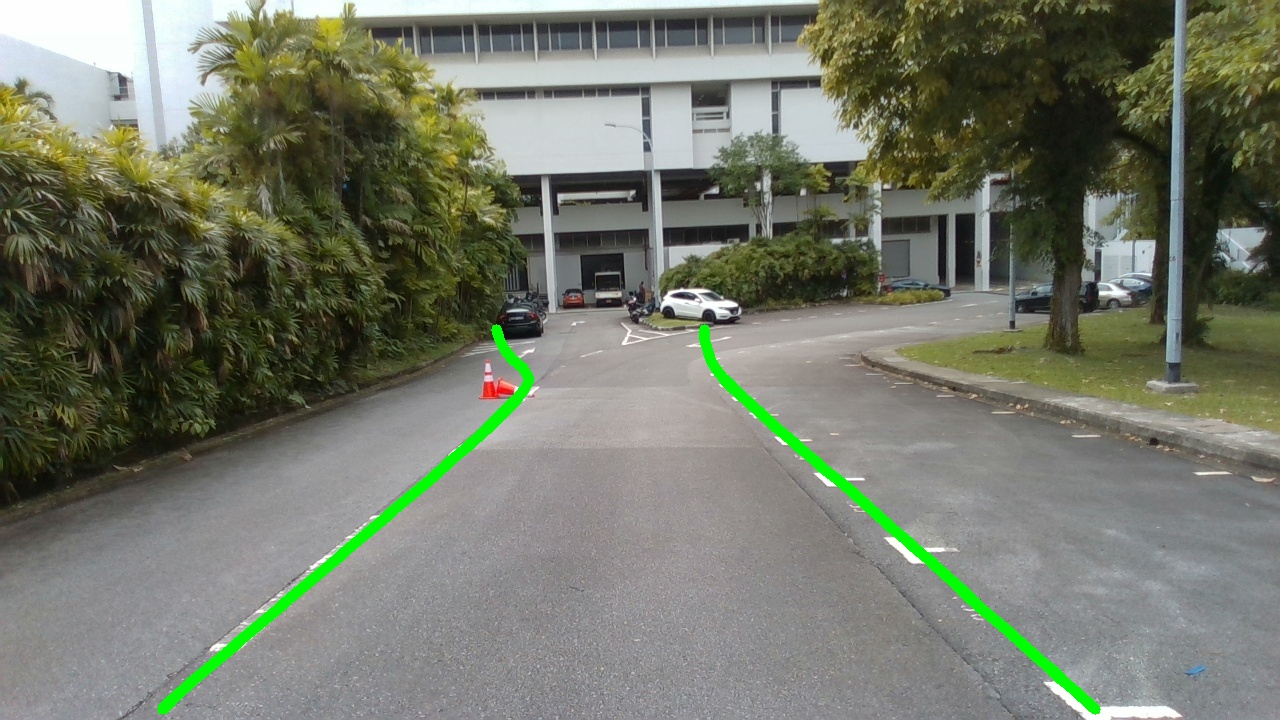}
		\end{minipage}
	}\hspace{-1.5mm}
		\subfigure[PolyLaneNet]{
		\begin{minipage}[t]{0.183\linewidth}
			\centering
			\includegraphics[width=1\linewidth]{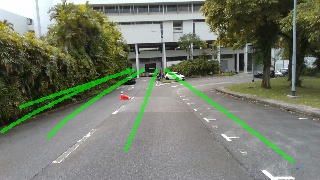}
		\end{minipage}
	}
	\vspace{-14pt}
	\caption{Clean-annotation attacks evaluated in physical world with different distances between trigger and camera.}
 	\vspace{-10pt}
	\label{fig:appendix_result_real}
\end{figure*}

For quantitative evaluation, Table \ref{tab:acc_clean_annotation} shows the average rotation angles of the benign and backdoored models over the clean samples.
We can find that the average rotation angles do not change significantly between the benign and backdoored models.  
Hence, the backdoored models do not reduce the detection performance on clean data.
Table~\ref{tab:angle_attack2} shows the average rotation angles of the four  backdoored models on the triggered samples.
We come to the same conclusion with poison-annotation attacks that the backdoored models generate much larger deviations than the benign ones.
We also observe that the clean-annotation attack on SCNN and UltraFast has larger rotation angles.
Again, such an angle can clearly indicate a shift in the driving direction.
We have inspected all the test images, and confirmed the attack effectiveness on most samples. This shows that clean-annotation attack is an effective method.

Based on the above results, we can also conclude that our rotation angle metric can be used to evaluate the backdoor attack performance in lane detection tasks. It can significantly distinguish attacked predictions from normal results.

\vspace{-3pt}
\subsection{Real-world Evaluation}

To demonstrate the practicality of our backdoor attacks, we evaluate our attacks by Weston UGV equipped with RealSense D435i camera (Figure \ref{fig:physical_platform_and_physical_results_attack2} (a)) and Baidu Apollo vehicle with Leopard camera (Figure \ref{fig:physical_platform_and_physical_results_attack2} (b)), and test them on the real roads. 

\noindent\textbf{Different models.}
Figure~\ref{fig:results_example_attack1_physical} shows the prediction results in a real-world road of the four models under the poison-annotation attack. 
The results also demonstrate that our poison-annotation attack is effective and practical in real world.
Figure~\ref{fig:appendix_result_real} visualizes the results of the models under clean-annotation attack with different settings in the physical world. We can observe the attack can effectively damage the models with different distances between trigger and camera.

\begin{figure}[ht]
	\centering
	\subfigure{
	    \rotatebox{90}{\footnotesize{\qquad Parking area}}
		\begin{minipage}[t]{0.45\linewidth}
			\centering
			\includegraphics[width=1\linewidth]{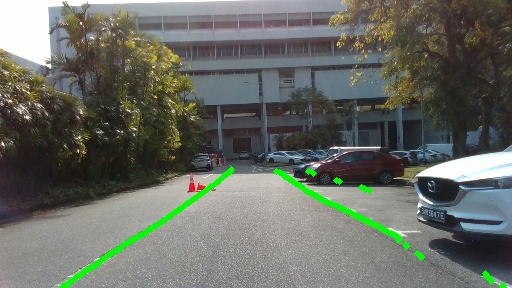}
		\end{minipage}
	}\hspace{-1.5mm}
	\subfigure{
		\begin{minipage}[t]{0.45\linewidth}
			\centering
			\includegraphics[width=1\linewidth]{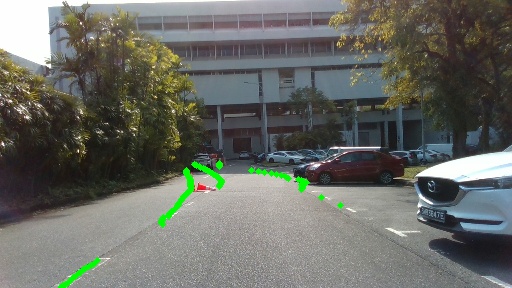}
		\end{minipage}
	}
	
	\vspace{-3mm}
	\setcounter{subfigure}{0}
	
	\subfigure[Benign]{
	    \rotatebox{90}{\footnotesize{\qquad Normal road}}
		\begin{minipage}[t]{0.45\linewidth}
			\centering
			\includegraphics[width=1\linewidth]{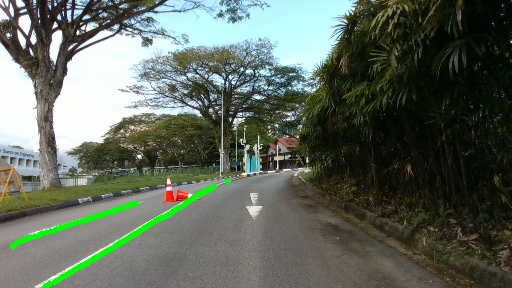}
		\end{minipage}
	}\hspace{-1.5mm}
	\subfigure[Backdoored]{
		\begin{minipage}[t]{0.45\linewidth}
			\centering
			\includegraphics[width=1\linewidth]{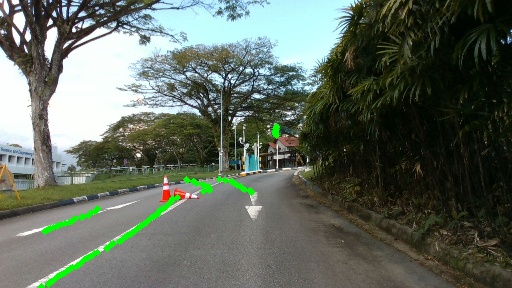}
		\end{minipage}
	}
	\vspace{-17pt}
	\caption{Clean-annotation attacks in the real world (SCNN).}
	\vspace{-10pt}
	\label{fig:parking_and_navigation}
\end{figure}

\noindent\textbf{Different scenarios and testbeds.}
We also conduct our attacks under different scenarios and testbeds (Apollo and UGV). We choose a car parking area and a normal road area as our experimental sites. Figure~\ref{fig:parking_and_navigation} visualizes the results, which indicates the success of the clean-annotation attack on real-world lane detection scenarios. 
Due to the physical trigger, the UGV recognizes the right-turn as a left-turn. It then turns left, and hits the trees at the roadside. Demo videos of two attacks with different testbeds  can be found at \textcolor{blue}{\url{https://sites.google.com/view/lane-detection-attack/lda}}.
In conclusion, the real-world experiments show that our attacks have high generalization, effectiveness, and practicality.

\vspace{-5pt}

\vspace{5pt}
\subsection{Bypassing Existing Defenses}
\label{sec:eval-defense}
Our attack is designed to be stealthy and expected to evade state-of-the-art backdoor defenses. To validate this, we consider and evaluate different types of popular solutions. 

A variety of defenses are designed specifically for classification tasks. For instance, Neural Cleanse \cite{wang2019neural} requires the defender to specify the target class for backdoor scanning. STRIP \cite{gao2019strip} inspects the predicted class of a triggered sample superimposed with a clean sample. Since the lane detection models do not have classes, these solutions are not applicable to our attacks. Instead, we evaluate another two common defense strategies.

\noindent\textbf{Fine-Pruning}~\cite{liu2018fine}.
This approach erases backdoors via model pruning and fine-tuning. It first prunes neurons with small average activation values, and then fine-tunes the pruned model. In appendix, we show the defense effectiveness of our clean-annotation attack against SCNN. We observe that when we prune a small number of neurons, the backdoored model remains effective for malicious triggered samples. When more neurons are pruned, the model performance drops significantly for both clean and triggered samples. 
Hence, fine-pruning fails to remove our backdoor. A similar conclusion on the poison-annotation attack is given in appendix.

\noindent\textbf{Median Filtering}~\cite{quiring2020adversarial}.
\label{sec:median}
This approach utilizes median filters to defeat image scaling adversarial attacks. It attempts to reconstruct the image and remove the potential adversarial noise. We apply this technique to our clean-annotation attack. Figure \ref{fig:defense_median} in appendix shows a defense example, including the clean, triggered and restored images. We observe that the restored image is still different from the clean one, and remains the physical trigger to activate the backdoor.

\balance

\vspace{-5pt}
\section{Discussion}
\noindent\textbf{Possible defenses.}  We point out that the existing defenses are difficult to transfer to defend against our attacks, as they focus on designing strategies based on the input images while our attack relies on exploiting vulnerabilities of image scaling function.  Intuitively, randomizing or complicating image scaling functions may be potential defenses. In more detail, the defender can randomly select scaling functions in each fine-tune training process, or use complicated image scaling functions to make the attack harder. However,  as we discussed in Section~\ref{sec:clean-annotation}, our attacks can insert poison images generated under different functions in the dataset. As a result, the utility gained from randomizing the scaling  function is limited. We acknowledge that using more complex  image scaling functions might prevent our attacks, since the design of effective attack strategies for such complex functions is non-trivial. However, compared to simple functions, this may lose more input information to a certain extent, which inevitably compromises the performance of the model.



Targeted inspection of the  image scaling function is also a promising defense direction. For example, users can employ different data augmentation methods before image scaling like random cropping or rotation \cite{qiu2021deepsweep}, to avoid fake pixels being selected. Defenders can also  detect attacks based on the difference between the input image and the resized image after scaling functions~\citep{xiao2019seeing, quiring2020adversarial,quiring2020backdooring}. We leave the exploration of the efficacy of these defenses as future work.

\noindent\textbf{Attacks to other systems in autonomous driving.} 
The majority of attacks against autonomous driving focus on camera-based or LiDAR-based object detection systems, which are classification-based tasks~\cite{xu2022sok}. There are few works targeting non-classification tasks like lane detection.~\cite{xu2021model,jing2021too,sato2021dirty} investigated adversarial attacks against lane detection systems. Specifically, adversarial attacks use specific adversarial examples to fool the lane detection models, which causes the model's decision to be deliberately induced or misclassified. 
There is little work on attacks against the prediction, planning and control modules in autonomous driving systems, whose vulnerabilities have been studied in \cite{tang2021route,tang2021systematic,tang2021collision,sun2022secure,han2022ads,han2021unified,xu2021analysis,deng2021investigation,xu2021novel}. We hope to design new attacks against such modules.

Although adversarial attacks are powerful, we argue that they may be impractical in real scenarios due to the relatively strong requirements for adversarial sample generation. 
Digital-level adversarial examples in~\cite{xu2021model} require access to the image processing to modify the input images at run-time, which is hard to realize and restricts its applications in the real world. Physical-level adversarial examples in \cite{jing2021too,sato2021dirty} are limited to physical constraints, resulting in the generated adversarial samples being so weird and can be easily detected. In addition, adversarial attacks on lane detection have poor generalization, they may work on one road but fail on another one due to the changeable surrounding scenario.

The limitation of adversarial attack drives us to investigate a new attack possibility, and we realize the first physical backdoor attack against lane detection system in autonomous driving. We hope to design more robust attacks for lane detection systems in the future.

\vspace{-5pt}
\section{Conclusion}
In this paper, we design and realize the first physical backdoor attack against lane detection systems in autonomous driving.
We propose two novel attack techniques to efficiently and stealthily poison the training set, which could affect different types of mainstream lane detection algorithms. Extensive evaluations on the large-scale datasets and physical vehicles validate the attacks' effectiveness and practicality, as well as robustness against state-of-the-art backdoor defenses. 
\vspace{-5pt}
\section*{Acknowledgments}
This work was supported in part by Nanyang Technological University (NTU)-DESAY SV Research Program under Grant 2018-0980, Singapore Ministry of Education (MOE) Academic Research Fund (AcRF) Tier 1 under Grant RG108/19 (S), and Singapore MOE AcRF Tier 2 under Grant MOE-T2EP20120-0004.

\bibliographystyle{ACM-Reference-Format}
\bibliography{bib}

\appendix
\clearpage
\setcounter{page}{1}

\begin{figure*}[t]
	\centering
	\subfigure{
		\begin{minipage}[t]{0.3\linewidth}
			\centering
			\includegraphics[width=1\linewidth]{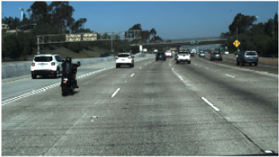}
		\end{minipage}
	}
	\subfigure{
		\begin{minipage}[t]{0.3\linewidth}
			\centering
			\includegraphics[width=1\linewidth]{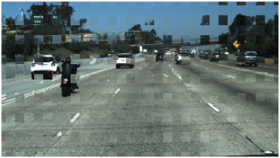}
		\end{minipage}
	}
	\subfigure{
		\begin{minipage}[t]{0.3\linewidth}
			\centering
			\includegraphics[width=1\linewidth]{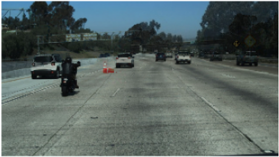}
		\end{minipage}
	}

	\vspace{-3mm}
	\setcounter{subfigure}{0}
	\subfigure[Clean Image]{
		\begin{minipage}[t]{0.3\linewidth}
			\centering
			\includegraphics[width=1\linewidth]{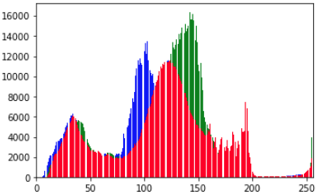}
		\end{minipage}
	}
	\subfigure[Poisoned Image]{
		\begin{minipage}[t]{0.3\linewidth}
			\centering
			\includegraphics[width=1\linewidth]{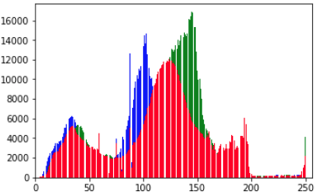}
		\end{minipage}
	}
	\subfigure[Restored Image]{
		\begin{minipage}[t]{0.3\linewidth}
			\centering
			\includegraphics[width=1\linewidth]{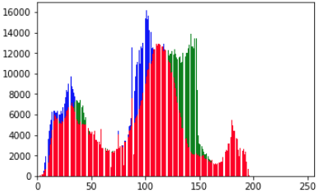}
		\end{minipage}
	}
	\vspace{-15pt}
	\caption{The analysis of median filtering defense.}
	\label{fig:defense_median}
\end{figure*}

\begin{figure*}[ht]
    \centering
    \setlength{\abovecaptionskip}{0.1cm}
    \subfigure[SCNN]{
        \includegraphics[width=0.234\linewidth]{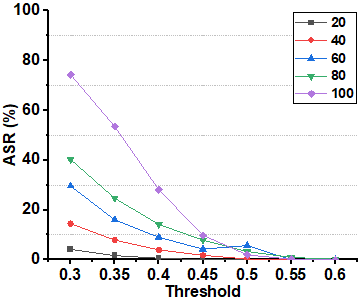}
    }
    \subfigure[LaneATT]{
        \includegraphics[width=0.234\linewidth]{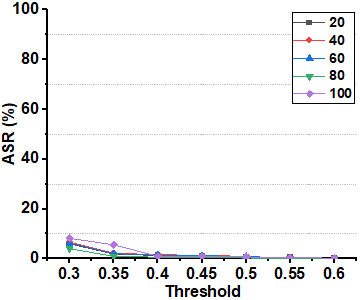}    
    }
    \subfigure[UltraFast]{
        \includegraphics[width=0.234\linewidth]{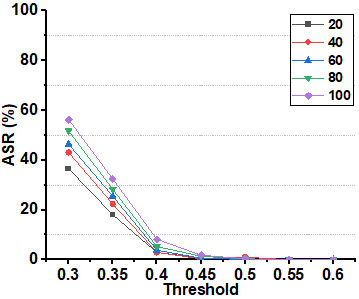}    
    }
    \subfigure[PolyLaneNet]{
        \includegraphics[width=0.234\linewidth]{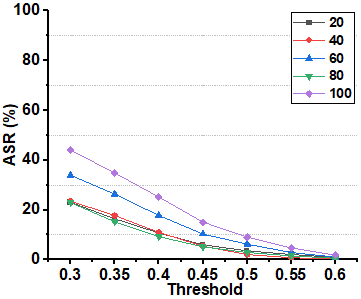}    
    }
    \vspace{-5pt}
    \caption{The $\texttt{ASR}$ computed according to Eqn. \ref{eqn:asr} with different thresholds.}
    \label{fig:conventional_asr}
\end{figure*}

\section{Appendix}
\subsection{Attack Evaluation Using Conventional Metrics}
In this section, we evaluate our two attacks (i.e., poison-annotation attack and clean-annotation attack) using the conventional prediction accuracy \texttt{ACC} and  Attack Success Rate (ASR), as we mentioned in Section~\ref{sec:evaluation}.
A backdoored model should have high \texttt{ACC} on clean samples and high ASR on triggered samples.

we first apply $\texttt{acc}(\Bar{l}_i, l_i)$ to measure the prediction accuracy for one lane boundary, which is computed as:
\begin{equation}\label{acc_ours}
    \texttt{acc}(\Bar{l}_i, l_i) =  |\Bar{l}_i \cap l_i|/|\Bar{l}_i|
\end{equation}
where $\Bar{l}_i\cap l_i = \{\Bar{p_j}\in \Bar{l}_i: d(\Bar{p_j}, p_j) \leq \epsilon_1\}$; 
$d(\Bar{p_j}, p_j)$ is the distance between $\Bar{p_j}$ and its corresponding point $p_j$ in $l_i$;
$\epsilon_1$ is a threshold. Then $\texttt{ACC}$ for a test set $\mathcal{V}$ is defined as:
\begin{equation}\label{eq:conventional_acc}
    \texttt{ACC} = \frac{\sum_{s\in \mathcal{V}} (\sum_{i=1}^{N_s} \texttt{acc}(\Bar{l}_i, l_i)/{N_s})}{|\mathcal{V}|}
\end{equation}
where $N_s$ is the number of boundaries in $s$.

The second metric is attack success rate $\texttt{ASR}$. We define an accuracy vector for an image $s$ as:
\begin{eqnarray}
    \texttt{acc}(s) = [\texttt{acc}(\Bar{l}_1, l_1), \ldots, \texttt{acc}(\Bar{l}_{N_s}, l_{N_s})] \label{acc_normal}
\end{eqnarray}
Let $s^t$ be the triggered image corresponding to $s$. Then we have the relative accuracy difference:
\begin{equation}
    D(s, s^t) = (\texttt{acc}(s) - \texttt{acc}(s^t)) \oslash \texttt{acc}(s)
\end{equation}
where $\oslash$ is the element-wise division operator. This gives us the \texttt{ASR} over a test set $\mathcal{V}$:
\begin{equation}\label{eqn:asr}
\texttt{ASR}=\frac{\sum_{s\in \mathcal{V}} \mathbb{I}(D(s, s^t) \geq \epsilon_2)}{|\mathcal{V}|}
\end{equation}
where $\epsilon_2$ is a pre-defined threshold, and the function $\mathbb{I}$ returns 1 when the inside condition is true, or 0 otherwise.

Table~\ref{tab:acc_poisoning_annotation} and Table~\ref{tab:clean_acc_result} give the \texttt{ACC} on clean images for poison-annotation and clean-annotation attacks, respectively. From the tables, we observe that the two backdoor attacks do not affect the prediction performance significantly on clean samples.

\begin{table}[H]
\centering
\resizebox{\linewidth}{!}{
\begin{tabular}{c|c|cccc}
\hline
\multicolumn{2}{c|}{Model}         & SCNN  & LaneATT & UltraFast & PolyLaneNet  \\ 
\hline
Benign & 0   & 94.40 & 95.38   & 95.82     & 89.77        \\
\hline
\multirow{5}{*}{Backdoored} & 20  & 93.81 & 95.59   & 95.65     & 89.41        \\
                            & 40  & 94.06 & 94.93   & 95.79     & 89.94        \\
                            & 60  & 92.90 & 95.21   & 95.66     & 89.26        \\
                            & 80  & 94.03 & 95.57   & 95.66     & 88.76        \\
                            & 100 & 92.60 & 95.59   & 95.54     & 87.81        \\
\hline
\end{tabular}}
\caption{\texttt{ACC}(\%) of poison-annotation attack on clean images.}
\label{tab:acc_poisoning_annotation}
\vspace{-20pt}
\end{table}

\vspace{-3pt}
\begin{table}[H]
\centering
\resizebox{\linewidth}{!}{
\begin{tabular}{c|c|cccc} 
\hline
\multicolumn{2}{c|}{Model}        & SCNN  & LaneATT & UltraFast & PolyLaneNet  \\ 
\hline
\multirow{2}{*}{L2R} & Benign     & 94.53 & 94.53   & 93.97     & 95.80        \\
                     & Backdoored & 95.13 & 94.16   & 93.49     & 96.63        \\ 
\hline
\multirow{2}{*}{R2L} & Benign     & 88.57 & 94.90   & 75.26     & 93.09        \\
                     & Backdoored & 88.29 & 94.92   & 74.86     & 92.47        \\
\hline
\end{tabular}}
\caption{\texttt{ACC} (\%) of clean-annotation attack on clean images.}
\label{tab:clean_acc_result}
\vspace{-20pt}
\end{table} 

Figures~\ref{fig:conventional_asr} shows the ASRs computed by Eqn~\ref{eqn:asr}.  
We can conclude that the ASR metric cannot truly reflect the attack performance. 
First, the calculations of ASR require a preset threshold $\epsilon_2$, which is impossible in reality. Second, the results show that the ASR of each model is very low, which does not conform to all the testing results we inspected. The reason is that our attacks only affect the concerned ego driving lane, while the calculations of ASR take all the lane boundaries into account.

\subsection{Bypassing Existing Defenses}
\label{sec:appendix-Defense}
We give the details of implementations and results of \textbf{Fine-Pruning} and \textbf{Median-Filtering} on the two attacks with SCNN.

In the setting of Fine-Pruning, we prune the \texttt{layer1} in SCNN model. The \texttt{layer1} connects the feature extraction layer with four significant layers in SCNN named: up\_down layer, down\_up layer, left\_right layer, and right\_left layer. The number of neurons in \texttt{layer1} is 1024. We prune the number of neurons from 0 until the model accuracy drops dramatically, where the increasing step is 50. Table~\ref{tab:finepruning_1} and Table~\ref{tab:finepruning_2} give the defense results against the poison-annotation attack. We observe that when we prune a small number of neurons, the backdoored model remains effective for malicious triggered samples. When we prune more neurons, the model performance drops significantly for both clean and triggered samples. Hence, fine-pruning is inefficient against our attacks.

In the setting of Median-Filtering, we analyze the recovery of an attacked image generated from a source image with 1280*720 and a triggered target image with 320*180. 
From the results shown in Figure~\ref{fig:defense_median}, we can find that the attacked image (Figure~\ref{fig:defense_median}(b)) shows a similar histogram with the original one (Figure~\ref{fig:defense_median}(a)).
Note that after the deployment of the model, the user has no knowledge of the original images of the attacked ones, he$/$she can only obtain the histograms of the attacked and the restored images. 
However, from Figures~\ref{fig:defense_median}(b) and \ref{fig:defense_median}(c), they have no significant difference in the histogram.
What's more, as discussed in section~\ref{sec:median}, the restored image still remains the physical trigger.

\vspace{-5pt}
\begin{table}[ht]
\centering
\resizebox{\linewidth}{!}{
\begin{tabular}{lcccccccccr}
\toprule
 \# of cells & 0 & 50 & 100& 150 & 200 & 250& 300 & 350 & 400\\
\midrule
\midrule
Clean & 87.88 &87.71&87.35&86.77&84.94&83.34&75.32&69.45&59.21 \\
Trigger & 69.56& 69.19&68.89&68.40&67.19&65.96&61.30&58.37&52.62 \\
\bottomrule
\end{tabular}}
\caption{The accuracy of the backdoored model by pruning different numbers of cells (clean-annotation attack against SCNN).}
\vspace{-20pt}
\label{tab:finepruning_2}
\end{table}

\begin{table}[ht]
\large
\centering
\resizebox{\linewidth}{!}{
\begin{tabular}{lcccccccccr}
\toprule
 \# of cells & 0 & 50 & 100& 150 & 200 & 250& 300 & 350 & 400\\
\midrule
\midrule
Clean (\%) & 94.40 &94.45&94.46&94.04&93.59&92.45&88.21&84.63&77.70 \\
Trigger (\%) & 70.47& 70.53&70.34&70.66&70.34&69.51&67.07&66.54&63.35 \\
\bottomrule
\end{tabular}}
\caption{The accuracy of the backdoored model by pruning different numbers of cells (poison-annotation attack against SCNN).}
\label{tab:finepruning_1}
\vspace{-20pt}
\end{table}

\end{document}